\documentclass[runningheads]{llncs}
\usepackage[T1]{fontenc}
\usepackage{graphicx}
\usepackage{todonotes}
\usepackage{comment}

\usepackage{xcolor}
\usepackage{todonotes}

\usepackage{cite}

\usepackage{graphicx}
\usepackage{balance}

\usepackage{amsmath}
\usepackage{comment}

\begin{document}
%
%
%
%
%





\title{Dynamic Modality and View Selection for Multimodal Emotion Recognition with Missing Modalities}

\titlerunning{Dynamic Modality and View Selection for MER with Missing Modalities}


%
%

\author{Luciana Trinkaus Menon\inst{1}\and
Luiz Carlos Ribeiro Neduziak\inst{1}\and
Jean Paul Barddal\inst{1}\and
Alessandro Lameiras Koerich\inst{2}\and
Alceu de Souza Britto Jr.\inst{1}}
\authorrunning{Menon, L.T. et al.}

%
\institute{Graduate Program in Informatics (PPGIa), Pontifícia Universidade Católica do Paraná (PUCPR), Curitiba, PR, Brazil \and
École de Technologie Supérieure (ÉTS), Montreal, Canada}
\maketitle              

\begin{abstract}
The study of human emotions, traditionally a cornerstone in fields like psychology and neuroscience, has been profoundly impacted by the advent of artificial intelligence (AI). Multiple channels, such as speech (voice) and facial expressions (image), are crucial in understanding human emotions. However, AI's journey in multimodal emotion recognition (MER) is marked by substantial technical challenges. One significant hurdle is how AI models manage the absence of a particular modality – a frequent occurrence in real-world situations. This study's central focus is assessing the performance and resilience of two strategies when confronted with the lack of one modality: a novel multimodal dynamic modality and view selection, and a cross-attention mechanism. Results on the RECOLA dataset show that dynamic selection-based methods are a promising approach for MER. In the missing modalities scenarios, all dynamic selection-based methods outperformed the baseline. The study concludes by emphasizing the intricate interplay between audio and video modalities in emotion prediction, showcasing the adaptability of dynamic selection methods in handling missing modalities.




\end{abstract}


\section{Introduction}


The world we live in is multimodal. In this context, modality refers to how we perceive and interact with our environment. Such a concept is crucial in advancing artificial intelligence (AI), as it underlines the AI systems' need to understand and integrate these different modalities effectively. The study of the complementary relationships between modalities such as visual (video), auditory (audio), textual (text), and sensory signals (e.g., heart rate variability) is essential for developing more sophisticated and context-aware AI systems \cite{Baltrusaitis_Ahuja_Morency_2019,Liu2023mmfusion}. 



Most current multimodal approaches assume that multiple modalities are always available and they carry complementary information \cite{Liu2023mmfusion}. This perspective is crucial in understanding how combining these modalities can lead to a richer and more comprehensive interpretation of underlying data. Each modality – visual, audio, textual, or otherwise – is believed to contribute unique insights. When merged, these insights form enhanced representations that are often more informative and accurate than any modality could achieve alone \cite{Praveen_2023}.


However, the ideal scenario of having all expected modalities available for a given task is not always the case \cite{DMelloKory2015,Li2024,Zhu2023,Vazquez-Rodriguez2023,Lin2023}. In the context of multimodal AI systems, the absence or unavailability of one or more modalities can significantly impact their performance and decision-making capabilities. Therefore, addressing missing modalities in multimodal learning is critical, as it reflects the challenges of dealing with real-world data that may not always conform to ideal settings.

This work evaluates two possible strategies to model the interaction of modalities (video+audio) in the context of emotion recognition. The first is a novel approach based on dynamic selection across the multimodal space. The second employs a well-known strategy using a neural network with an attention-based mechanism to jointly learn the modalities, inspired by the method proposed in \cite{Praveen_2022}. In other words, we focus on assessing the impact of one modality's absence on model performance within these distinct multimodal AI approaches. Thus, two research questions guided our experiments. The first addresses the proposed dynamic selection method, as (RQ1) -- ``Could dynamic selection of modalities and views be a promising approach for a multimodal AI method?''. The second research question concerns the impact of a missing modality on each evaluated multimodal approach, formulated as (RQ2) -- ``What is the impact on emotion recognition performance when one modality (video or audio) is missing?''. To this end, we simulate the loss of a modality by replacing the corresponding features with zeros.


The contribution of this paper is twofold: (i) it introduces a novel multimodal method for emotion recognition based on dynamic modality and view selection, and (ii) it assesses how different approaches (dynamic selection vs. attention mechanism) respond to the absence of a specific modality. This work scrutinizes the challenges faced by AI in multimodal emotion recognition (MER). Also, it explores innovative methodologies, aiming to push the boundaries of what these sophisticated models can achieve in understanding the complex landscape of human emotions.

The remainder of this paper is structured into six sections. Section 2 discusses related works on MER and missing modalities. Section 3 outlines the proposed method for dynamic modality and view selection. Section 4 details the strategies employed to assess the impact of missing modalities on the proposed method and the baseline. Section 5 presents the experimental results and corresponding analysis. Finally, our conclusions and avenues for further research are presented in Section 6.

\section{Related Works}


The field of MER has been the subject of extensive research, given its applicability in various domains \cite{Praveen_2023,cardinal2015ets,ortega2018multimodal,ortega2019emotion,Praveen_2022,Aruna2023,aslam2023privileged,Liu2023mmfusion,Cheng2024}. This section highlights relevant works addressing the importance of emotion recognition through multiple modalities and dealing with missing modalities during inference.





Multiple modalities are essential in emotion recognition because humans simultaneously express emotions through various channels. Human communication encompasses both verbal and non-verbal expressions, highlighting our dependence on multimodal cues. When pronounced with emphasis in a deep voice, a simple word can convey a more intense emotion to the listener -- a nuance that eludes capture through the analysis of linguistic features alone. Complex emotions, such as fear, are more readily recognized through the interplay of facial expressions and variations in pitch than relying solely on the transcripts of spoken words \cite{Aruna2023}.

Relying on a single modality may lead to incomplete or inaccurate assessments of emotional states. In contrast, multimodal models are shown to enhance reliability and accuracy when compared to unimodal ones \cite{Aruna2023,Liu2023mmfusion}. However, multimodal data are heterogeneous and complex. Some central technical challenges of using multimodal data in machine learning include choosing the most suitable data representations, addressing misalignment between elements of different modalities, and managing intramodal and cross-modal data correlation and fusion approaches \cite{Aruna2023}. Dealing with missing data is one of the most critical challenges in multimodal fields ~\cite{Liu2023mmfusion}.


Several factors can contribute to the absence or unavailability of specific modalities in MER approaches at inference time \cite{aslam2023privileged,Lin2023,Cheng2024}, such as malfunction of cameras or microphones resulting in the inability to capture specific modalities, disabling or restricting certain sensors due to privacy concerns, poor lighting conditions can hinder the accurate detection of facial expressions, background noise that may affect the capture and interpretation of vocal expressions, invasiveness of sensor to capture physiological signals, etc. Understanding these causes is essential for developing strategies to handle missing modalities effectively. 


Handling missing modalities in multimodal approaches often involves three strategies to adapt to such situations \cite{aslam2023privileged,Lin2023,Lin2023a,Cheng2024,Vazquez-Rodriguez2023}: (i) imputing or filling in missing modalities using data imputation or leveraging information from other available modalities; (ii) designing models that can gracefully handle scenarios where certain modalities are missing, potentially by learning to rely more on available modalities; (iii) developing models that can adapt to varying modalities or different data distributions, allowing for better generalization when faced with missing modalities.


Several works propose innovative approaches to the absence of modalities during inference. When dealing with missing modalities, common approaches often perform imputations to address the absence of modalities before proceeding with additional computations. Simple imputation methods, such as filling missing values with zeros, are straightforward but may lead to considerable inaccuracies.

Other noteworthy contributions include the work of Da Silva–Filarder et al.~\cite{DaSilvaFilarder2021}, which studies multimodal variational autoencoders and states that a critical property of multimodal generative models is to have efficient approaches to deal with missing modalities and to enable cross-generation. Cross-generation involves using a subset of input modalities to generate output modalities missing from the input subset. The authors propose two methods: latent component dropout (LCD) and exhaustive cross-generation (ECG). LCD is based on randomization and simulates missing modalities by applying a dropout mask for each modality to individual elements of the latent variables. It then reverts to a prior expert for those elements the mask selects. ECG is a method based on brute force that considers all possible subsets of modalities.


Li et al.~\cite{Li2024} also address the issue of missing modalities. According to the authors, these missing data harm extracting features of multimodal data, resulting in a decline in model performance and inaccurate results. Therefore, a multiple multi-head attention network based on an encoder with missing modalities is proposed. The multi-head attention represents the modality based on the entire sequence, and the cross-mapping is used to obtain the relationship between the modalities. Random missing modalities are encoded and combined with an optimization module to enhance the association between missing and non-missing modalities. The encoder and decoder module obtain global information and map global information to multiple spaces.


Zhu et al.~\cite{Zhu2023} proposed an invariant feature for a missing modality imagination network. This method relies on two encoders: the specificity encoder, responsible for extracting high-level features from raw input features, and the invariance encoder, which takes the modality-specific features obtained by the specificity encoder as input to extract modality-invariant features. Additionally, the method incorporates an invariant feature-based imagination module that predicts the modality-specific features of the missing modality based on the available modality and generates a joint representation.


Vazquez-Rodriguez et al.~\cite{Vazquez-Rodriguez2023} proposed a transformer-based architecture for continuous prediction of arousal and valence, even with missing input modalities. A multimodal transformer is used as an encoder to obtain representations from the different modalities, and a transformer decoder is used to process those representations and make predictions. An encoder-decoder attention mechanism (cross-attention) of the transformer decoder is used to weigh the importance of different modalities. The transformer decoder is auto-regressive, considering past predicted values when doing the current inference, which is essential when performing time-continuous predictions.


Other interesting works present computational multimodal frameworks based on transformer architecturea and attention mechanisms for MER with incomplete data ~\cite{Lin2023,Cheng2024,Liu2024,Nguyen2023}. Despite the impressive performance when a massive dataset is available, such an approach involves more complex operations, making interpretation less intuitive. In addition,  attention mechanisms and Transformers can become computationally costly, even when some modalities are absent, due to the attention structure that may involve all pairs of elements in the sequence.

Dynamic selection can offer specific advantages compared to attention mechanisms and transformer architecture when dealing with missing modalities in multimodal approaches. Dynamic selection-based methods provide better interpretability, as the choice of specific regressors/classifiers in the absence of modalities can be easily analyzed. Additionally, dynamic selection can be more computationally efficient than the attention structure, which may involve all sequence elements. Finally, dynamic selection is less dependent on large datasets for training, as observed in our experiments.











Each approach has advantages and challenges, and the choice should be based on the specific demands of the task. In this work, we will analyze the impact of missing modalities in solutions based on dynamic selection and attention mechanisms in the context of arousal-valence regression. Can more straightforward solutions based on the imputation method of filling missing values with zeros effectively address the problem of absent modalities?


\section{Proposed Dynamic Modality and View Selection Method}

\begin{figure}[]
\centering
\includegraphics[width=\textwidth]
{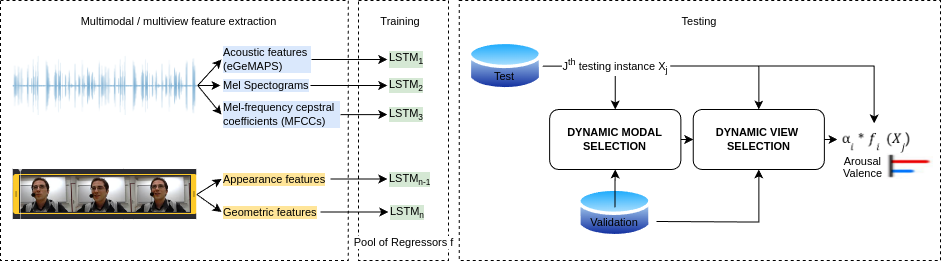}
\caption{
Dynamic modality and view selection method. The audio features include acoustic features, MFCCs, and Mel spectrograms. The video features include appearance features and geometric features. All regressors are trained separately, and a pool of regressors is obtained. The best modal is selected in the dynamic modal selection phase, after that, in the intra-modal dynamic view selection each model receives a weight to evaluate each test case according to its assertiveness in the competence zone.
}
\label{fig:figure0}
\end{figure}

The proposed dynamic selection-based method considers features from AVEC'16 \cite{AVEC2016}, encompassing audio and video modalities. As shown in Fig.~\ref{fig:figure0}, the audio features include acoustic features, MFCCs, and Mel spectrograms. On the video side, we incorporate appearance features and geometric features. Each regressor is trained independently, resulting in a pool of regressors denoted as $F = {f_{1}, f_{2}, \dots, f_{N}}$, where $N$ is the total number of regressors.

Two LSTM layers with 256 cells each are employed to consider the temporal structure of the data. For the recurrent layers, the input is segmented into sequences of 6 seconds, corresponding to 150 time steps (frames) at a sampling rate of 16kHz. The motivation for using a simple 2-layer LSTM model is to evaluate how promising dynamic selection is considering weak regressors.


Initially, the LSTM models are evaluated in the dynamic modal selection (DMS) phase. DMS is performed by training a meta-classifier with concatenated outputs from each regressor (a vector of dimension N). Training is conducted on the validation set, and the ideal output is defined as the modal (audio or video) with the best mean Concordance Correlation Coefficient (CCC), the modal whose predictions are closest to the proper labeling. 

After the modal is selected, intra-modal dynamic view selection (DVS) is performed. Each model from the selected modal receives a weight to assess each test case $x_{j}$ based on its performance in the competence region $\Psi$ - set composed of the K-nearest neighbors of $x_{j}$ in the validation set. The DVS selects the regressor with the smallest accumulated error in the competence region or combines all the regressors or a subset using weighted averaging according to a calculated weight $\alpha_{i}$ of regressor $f_{i}$.

The impact of the absence of a modality was accessed using traditional dynamic selection techniques \cite{Moura_Cavalcanti_Oliveira_2021}, adapted here for the multimodal problem: dynamic selection (DS), dynamic weighting (DW) and dynamic weighting selection (DWS). 

In the \textbf{DS}, we select the regressor from the previously chosen modal with the smallest accumulated error in the competence region. \textbf{DW} combines all regressors from a pre-selected modal using weighted averaging. For each test pattern $x_{j}$, its competence region $\Psi$ is calculated. For each item in $\Psi$, a weight $d_{k}$ is calculated using Eq.~\ref{eq:1}, where ${dist_k}$ is a distance measure between the item in the competence region $t_{k} \in \Psi$ and the test pattern $x_{j}$. The vector $d_{1}, d_{2}, ..., d_{k}$ is used to calculate the weight $\alpha_{i}$ of regressor $f_{i}$ using Eq.~\ref{eq:2}, where N is the size of the selected modal regressors pool, $k$ represents the neighbor index, and $sqe_{k,i}$ is the squared error of regressor $i$ calculated using the item $t_{k} \in \Psi$. Finally, \textbf{DWS} combines a subset of regressors, and regressors with the accumulated
error in the upper half of the error interval  $E_{i} > (E_{max} -E_{max})/2$ are discarded. The method for calculating the weights of the regressors and the strategy for combining the models are the same as the \textbf{DW} algorithm (Eqs.~\ref{eq:1} and \ref{eq:2}).

\begin{equation} \label{eq:1}
d_{k} = \frac{(\textrm{dist}_k)^{-1}}{\sum_{j=1}^{K}{(\textrm{dist}_{j})^{-1}}}
\end{equation}

\begin{equation} \label{eq:2}
\alpha_{i} = \frac{\left[\sum_{k=1}^{K}{(d_{k} * sqe_{k,i})}\right]^{-1}}{\sum_{n=1}^{N}\left[\sum_{k=1}^{K}{(d_{k} * sqe_{k,i})}\right]^{-1}}
\end{equation}

It is important to emphasize that tests were conducted with the standard methods of dynamic selection, DS, DW, and DWS, with $K$ varying from $5$ to $150$. All results presented in this work are with $K = 100$. In addition, as a baseline, we compute the simple average of the regressors' outputs and utilize the cross-attention architecture proposed by ~\cite{Praveen_2022}. 

\section{Missing Modality Analysis}

The experimental approach was structured in three distinct phases. The first phase involved training the models using both 
 audio and video modalities. In the second phase, the audio modality was disabled, and the relative contribution of the video modality to the regression task was assessed. Finally, in the third phase, the video modality was disabled, allowing for the evaluation of the audio modality's relative contribution to the regression task.

This approach was inspired by sensitivity analysis methods used in \cite{NaikKiran,featureSelectionAlceu,VERIKAS2002}. In these studies, sensitivity analysis was conducted at the feature level, examining the impact of subsets of features on the overall performance of a machine learning model. In the present work, however, the sensitivity analysis was performed by generating a zero feature vector for the modality intended to be disabled. Subsequently, the model was tested with a fusion of this feature vector from the disabled modality and the active feature vector from the other modality.

Such analysis provided us valuable insights into how each modality independently influences the model's performance and how the strategies employed by dynamic selection and cross-attention handle the absence of specific modalities. By comparing the results from each phase, one could discern the individual and combined effects of audio and video modalities. Moreover, this analysis sheds light on the sensitivity of the proposed methods when confronted with missing modalities.

\section{Experiments}

The remote collaborative and affective interactions (RECOLA) dataset \cite{recola} represents an extensive source of multimodal data, encompassing extracted features and raw data from various modalities, including audio, video, and physiological recordings (electrocardiogram and electrodermal activity). The labeling of the first five minutes of interaction for 18 participants is available.


The data is labeled within the repository, adhering to a continuous emotional scale. This labeling is mapped into a two-dimensional space, a psychologically grounded method for describing emotions through the linear combination of arousal and valence. The concept of representing emotions in arousal and valence follows the circumplex model proposed by Russell \cite{russell1980circumplex}.

The official metric for evaluating the performance of the problem is the CCC~\cite{AVEC2016}, which captures the co-variation relationship between predictions and ground truth and accounts for any deviation. As a result, it offers a more accurate representation of the alignment between predictions and ground truth \cite{e25101440}. Higher CCC values signify excellent performance in terms of consistency and accuracy. The calculation process for CCC is as follows:

\begin{equation}
CCC = \frac{2 * \rho * \sigma_{y} * \sigma_{\hat{y}}}{\sigma_{y}^2 + \sigma_{\hat{y}}^{2} + (\mu_{y} - \mu_{\hat{y}})^{2}}
\end{equation}

\noindent where \(\rho\) is the Pearson correlation coefficient, \(\sigma_{y}\) and \(\sigma_{\hat{y}}\) are the standard deviations and \(\mu_{y}\) and \(\mu_{\hat{y}}\) are the means of actual and predicted emotional state.

This experiment emphasized two primary modalities: audio and video. The eGeMAPS acoustic feature set was employed for the audio component, which was extracted using the OpenSmile software and is available within the RECOLA dataset. Additionally, feature sets based on Mel-frequency cepstral coefficients (MFCCs) and Mel spectrograms, both of which were extracted by the authors of this work, were utilized. The video component, on the other hand, has been focused solely on extracted features from the RECOLA dataset, including geometric features derived from 49 distinct facial landmarks and appearance features obtained by a principal component analysis (PCA) from 50,000 LGBP-TOP features.

\subsection{Experimental Protocol}

An experimental protocol based on the k-fold cross-validation method has been implemented to ensure the robustness and reliability of our findings. 

The dataset comprised data from 18 individuals. To balance training and testing and ensure that our model was tested on unseen data, we allocated three individuals for testing and three for validation. The remaining participants were used for training. The experimental setup was repeated ten times, each time with a different configuration, to enhance the generalization of our results. In each iteration of the experiment, the participants were randomly shuffled. 



We deliberately introduced a modality-absent condition to simulate a real-world scenario. These simulations are essential for assessing the robustness and adaptability of our model under less-than-ideal conditions. To simulate the absence of a modality a zero vector was used. In practical terms, this meant that for any given instance where a particular modality was supposed to be missing, its feature values were replaced with zeros. This approach effectively mimics scenarios where a modality's data is entirely unavailable, allowing us to observe how the model performs when deprived of information from one of the modalities.


\subsection{Results}


This section offers an in-depth analysis of the outcomes achieved by employing the proposed techniques of dynamic modal and view selection and cross-attention mechanism under ideal conditions and on a modality-absent condition.

Table \ref{tab:table1} displays the arousal and valence results, in terms of CCC, for the pool of regressors $F$. The findings reveal that the audio modality better represents the arousal dimension, with acoustic features, MFCCs, and Mel spectrograms, achieving CCC values of $0.69$, $0.64$, and $0.68$. Conversely, valence is more accurately represented by the video modality, with its appearance features and geometric features representations, achieving CCC values of $0.48$ and $0.56$.

\begin{figure}[!ht]
\centering
\includegraphics[width=\textwidth]{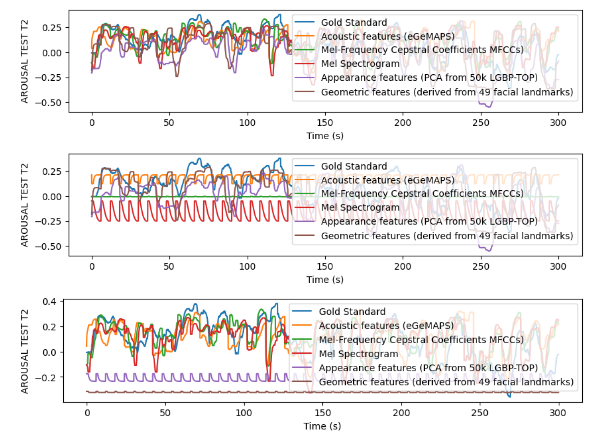}
\caption{
Arousal gold standard and prediction of models based on acoustic features, MFCCs, Mel spectrograms, appearance features, and geometric features. The image was generated using test case T2 (second person from the test set) of the second cross-validation fold (k=2). From top to bottom, we have (i) prediction with all active modalities, (ii) prediction with the absence of audio modality, and (iii) prediction with the absence of video modality.
}
\label{fig:figure1}
\end{figure}

Fig.~\ref{fig:figure1} shows the arousal prediction of all models for the same test case, considering scenarios where both modalities are available and when each modality is individually unavailable. Under ideal conditions, all models exhibit a consistent pattern with similar predictions. However, at certain moments, one model aligns more closely with the gold standard, while another performs better at other times. When a modality is absent, a noticeable decline in performance is observed among models relying on representations of that particular modality. 

For arousal, under ideal conditions, with all modalities available - video and audio, the highest performance with $CCC = 0.72$ was observed when employing DW, DWS, and a simple mean of all regressors' outputs. DW and DWS yielded the best outcomes in the valence dimension with $CCC = 0.54$ and $CCC = 0.53$, surpassing DS and mean of regressors' outputs, which registered $CCC = 0.46$. Cross-attention results included $CCC = 0.46$ for arousal and $CCC = 0.41$ for valence. Detailed results are shown in Table \ref{tab:table2}.

Regarding arousal, methods based on dynamic selection (DW and DWS, $CCC = 0.72$) outperformed the top-performing regressor alone, relying solely on acoustic features, $CCC = 0.69$. It shows that dynamic selection of modalities can be a promising approach for a multimodal AI method. Valence achieved its peak performance with geometric features, $CCC = 0.56$, and none of the proposed methods managed to surpass this benchmark in valence prediction.

\begin{table*}[!ht]
\caption{CCC for arousal and valence encompassing models based on acoustic features, MFCCs, Mel spectrograms, appearance features, and geometric features. Models were trained with two layers of LSTM with 256 cells and a time window of 6 seconds.}
\centering
\label{tab:table1}
\begin{tabular}{|c|c|c|}
\hline
\textbf {Features} & \textbf{Arousal} & \textbf{Valence} \\ \hline
Acoustic            & \textbf{0.69±0.06} & 0.18±0.07 \\ \hline
MFCCs               & 0.64±0.06 & 0.35±0.08 \\ \hline
Mel Spectrograms    & 0.68±0.06 & 0.22±0.09 \\ \hline
Appearance          & 0.42±0.09 & 0.48±0.06 \\ \hline
Geometric           & 0.41±0.09 & \textbf{0.56±0.14} \\ \hline\end{tabular}
\end{table*}

\begin{table*}[!ht]
\caption{CCC for arousal and valence encompassing the mean of the regressors' outputs, dynamic selection (DS, DW, DWS), and cross-attention-based methods. The DS, DW, and DWS results were generated with K = 100. 
}
\centering
\label{tab:table2}
\begin{tabular}{|c|c|c|}
\hline
\textbf{Approach} & \textbf{Arousal} & \textbf{Valence} \\ \hline
DS              & 0.67±0.06   &      0.46±0.08       \\ \hline
DW              & \textbf{0.72±0.04}   &      \textbf{0.54±0.10}       \\ \hline
DWS             & \textbf{0.72±0.04}   &      0.53±0.10       \\ \hline
Cross-Attention & 0.46±0.13 & 0.41±0.17 \\ \hline
Mean            & \textbf{0.72±0.04}   &      0.46±0.08       \\\hline
\end{tabular}
\end{table*}

\subsection{Impact of Missing Modalities}

\begin{figure}[!ht]
\centering
\includegraphics[width=\textwidth]{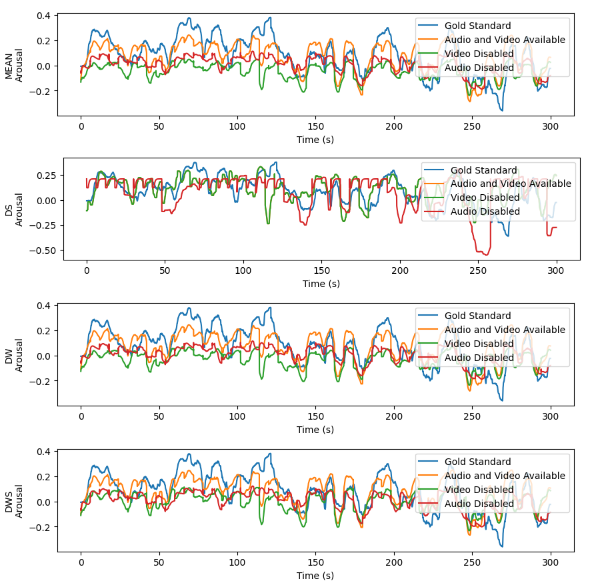}
\caption{Comparison of arousal gold standard, prediction with all active modalities, prediction with the absence of audio modality, and prediction with the absence of video modality of the mean of the regressors' outputs and dynamic selection-based methods (DS, DW, DWS). The image was generated using test case T2 (second person from the test set) of the second cross-validation fold (k=2).
}
\label{fig:figure2}
\end{figure}

Our second research question is related to how the different approaches (dynamic selection and cross-attention) respond to the absence of a specific modality. Tables \ref{tab:table3} and \ref{tab:table4} display the arousal and valence results, in terms of CCC, of all comparison methods - encompassing mean of the regressors' outputs, dynamic selection (DS, DW, DWS), and cross-attention. Fig.~\ref{fig:figure2} compares the arousal gold standard, prediction with all active modalities, and with the absence of each modality.

\begin{table*}[!htp]
\caption{Arousal results, in terms of CCC, encompassing the mean of the regressors' outputs, dynamic selection (DS, DW, DWS), and cross-attention-based methods. The results are presented in the following scenarios: audio (A) and video (V) available, audio disabled and video disabled, simulating the absence of a modality with a zero vector.}
\centering
\label{tab:table3}
\begin{tabular}{|c|c|c|c|c|c|}
\hline
\textbf{Modalities}       & \textbf{Mean} & \textbf{DS} & \textbf{DW} & \textbf{DWS} & \textbf{Cross-Attention} \\ \hline
A and V available & \textbf{0.72±0.04}     & 0.67±0.06   & \textbf{0.72±0.04}   & \textbf{0.72±0.04}    & 0.46±0.13                \\ \hline
V disabled  & 0.61±0.09 & 0.67±0.06 & \textbf{0.71±0.05} & \textbf{0.71±0.05} & 0.49±0.12 \\ \hline
A disabled  & 0.31±0.05 & 0.35±0.15 & \textbf{0.41±0.17} & 0.40±0.17 & 0.23±0.12 \\ \hline
\end{tabular}

\end{table*}

\begin{table*}[!htp]
\caption{Valence results, in terms of CCC, encompassing the mean of the regressors' outputs, dynamic selection (DS, DW, DWS), and cross-attention-based methods. The results are presented in the following scenarios: audio (A) and video (V) available, audio disabled and video disabled, simulating the absence of a modality with a zero vector.}
\centering
\label{tab:table4}
\begin{tabular}{|c|c|c|c|c|c|}
\hline
\textbf{Modalities}       & \textbf{Mean} & \textbf{DS} & \textbf{DW} & \textbf{DWS} & \textbf{Cross-Attention} \\ \hline
A and V available & 0.46±0.08 & 0.46±0.08 & \textbf{0.54±0.10}  & 0.53±0.10  & 0.41±0.17 \\ \hline
V disabled           & 0.27±0.08 & 0.28±0.18 & \textbf{0.30±0.17} & \textbf{0.30±0.17} & 0.13±0.15 \\ \hline
A disabled           & 0.32±0.06 & 0.49±0.09 & \textbf{0.59±0.09} & 0.56±0.09 & 0.40±0.14 \\ \hline
\end{tabular}
\end{table*}

In the context of arousal, the cross-attention method demonstrated heightened robustness in terms of sensitivity, exhibiting a $6.52\%$ increase in CCC when the video modality was absent, compared to the ideal scenario where both audio and video modalities were available. Contrarily, the remaining methods either sustained their performance or experienced some loss. Among the dynamic selection-based methods, DS exhibited no performance lowering, while DW and DWS showed a minimal decrease of $1.39\%$. The mean method observed the most substantial decline, recording a significant loss of $15.28\%$.

More pronounced performance losses were observed when the audio modality was absent. Several approaches witnessed a decline of over $50\%$ in performance, which is understandable as the audio modality most effectively represents the arousal dimension. DW emerged as the most robust approach in scenarios without audio, experiencing a performance decline of $43.06\%$ compared to the scenario with all available modalities. Following closely, DWS and DS demonstrated a CCC decline of $44.44\%$ and $47.76\%$.



A contrasting pattern was observed in valence, where disabling the audio modality yields superior results. When the video modality is absent, DS proves to be the least sensitive approach with a performance decline of $39.13\%$. In the same scenario, when the audio modality is missing, DW and DS emerged as the most robust methods, showcasing a $9.26\%$ and $6.52\%$ increase in CCC, respectively. 

In the cross-attention method, concerning arousal, there was a notable $6.52\%$ increase in CCC when the video modality was absent but a substantial $50\%$ decline in performance when the audio modality was disabled. Regarding valence, promising outcomes were observed when the video modality was turned off, with a performance lowering of $2.44\%$. However, turning off the video modality resulted in a significant decline of $68.29\%$ in performance compared to the ideal scenario where both audio and video modalities were available.

Considering the scenarios of missing modalities in arousal and valence dimensions, all dynamic selection-based methods (DS, DW, DWS) consistently outperformed the baselines, mean of all regressors' output, and cross-attention-based method. 

\subsection{Discussion}


The proposed approach performs better in arousal than valence, especially when the audio features are available. It may be related to the fact that arousal, which relates to the emotional intensity or activation level, might be more distinctly captured in tone of voice, volume, and speech rate, even without visual cues. For example, screams or high intonations may indicate a more excited emotional state. Elements like rhythm and timbre in speech also reflect emotional excitement; rapid rhythm or timbre changes can indicate more intense emotional states. Audio data carry significant information about the emotional state and can be quite effective in capturing the subtleties of arousal levels.

Auditory cues might be less effective in conveying valence levels. Valence, associated with the positivity or negativity of emotions, is often reflected in facial expressions and might be more nuanced and complex to discern from audio alone. Visual cues are critical in identifying the valence levels, making video a more informative modality for this dimension.

MER using dynamic modality and view selection appears to be an effective strategy for combining audio and video modalities, showing promising results under ideal conditions. The highest performances were observed when employing DW, with arousal $CCC = 0.72$ and valence $CCC = 0.54$.

Furthermore, dynamic modality and view selection techniques exhibit notable robustness when confronted with the absence of specific modalities. In scenarios where the audio modality was absent DW also demonstrated heightened robustness in terms of sensitivity, exhibiting a performance decline of $43.06\%$ in arousal and a increase of $9.26\%$ in valence, compared to the ideal scenario where both audio and video modalities were available. The cross-attention method emerged as the most robust approach when the video modality was absent, exhibiting a $6.52\%$ increase in CCC in arousal (DW demonstrated a decline of $1.39\%$). In valence dimension DS proves to be the least sensitive approach in scenarios without video with a performance decline of $39.13\%$ (DW demonstrated a decline of $44.44\%$).

For (RQ1) -- ``Could dynamic selection of modalities and views be a promising approach for a multimodal AI method?'', the results affirmatively show that dynamic selection-based methods are promising. However, the outcome of missing modalities revealed interesting nuances, addressing the research question (RQ2) -- ``What is the impact on emotion recognition performance when one modality (video or audio) is missing?'', several approaches witnessed a decline of over $50\%$ in performance when a modality is absent, emphasizing the importance of each modality in contributing to accurate predictions.

\section{Conclusion}

Our investigation into the representation of time-continuous emotions, particularly in arousal and valence dimensions, through different dynamic selection approaches has yielded valuable insights. Even under less-than-ideal conditions, MER systems have demonstrated their versatility and reliability.

The findings reveal that DW show the highest performance in arousal and valence predictions under ideal conditions, with both modalities available.  In the missing modalities scenarios, all dynamic selection-based methods (DS, DW, and DWS) outperformed the baselines, mean of all regressors' output, and cross-attention-based method. The study concludes by emphasizing the intricate interplay between audio and video modalities in emotion prediction, showcasing the adaptability of dynamic selection methods in handling missing modalities. 

Finally, it is important to highlight that we have employed simple two-layer LSTMs for all modalities. Replacing such a model and handcrafted feature representations by pre-trained vision transformers and large language models must improve the performance for each modality and the final ensemble.








\balance

%
%
%
\bibliographystyle{splncs04}
\bibliography{IEEEabrv}

\end{document}